\documentclass[10pt,conference]{IEEEtran}
\IEEEoverridecommandlockouts
\usepackage{cite}
\usepackage{amsmath,amssymb,amsfonts}
\usepackage{algorithmic}
\usepackage{graphicx}
\usepackage{textcomp}
\usepackage{xcolor}
\usepackage{pgfplots}
\pgfplotsset{width=10cm,compat=1.9}
\usepackage{array}
\usepackage{listings}
\usepackage[most]{tcolorbox}
\usepackage{balance}
\usepackage[nolist,nohyperlinks]{acronym}
\def\BibTeX{{\rm B\kern-.05em{\sc i\kern-.025em b}\kern-.08em
    T\kern-.1667em\lower.7ex\hbox{E}\kern-.125emX}}

\definecolor{nicolasred}{rgb}{0.7804, 0, 0.0431}
\definecolor{nicolascyan}{rgb}{0.1882, 0.7137, 0.7725}

\begin{document}

\title{Telecom Language Models: Must They Be Large?}

%\author{\IEEEauthorblockN{Nicola Piovesan}
%\IEEEauthorblockA{Paris Research Centre, Huawei %Technologies,\\ Boulogne-Billancourt, France\\
%nicola.piovesan@huawei.com}
%\IEEEauthorblockN{Antonio De Domenico}
%\IEEEauthorblockA{Paris Research Centre, Huawei %Technologies,\\ Boulogne-Billancourt, France\\
%antonio.de.domenico@huawei.com}
%\and
%\and

\author{\IEEEauthorblockN{Nicola Piovesan}
\IEEEauthorblockA{\textit{Paris Research Centre} \\
\textit{Huawei Technologies}\\
Boulogne-Billancourt, France \\
nicola.piovesan@huawei.com}
\and
\IEEEauthorblockN{Antonio De Domenicoo}
\IEEEauthorblockA{\textit{Paris Research Centre} \\
\textit{Huawei Technologies}\\
Boulogne-Billancourt, France \\
antonio.de.domenico@huawei.com}
\and
\IEEEauthorblockN{Fadhel Ayed}
\IEEEauthorblockA{\textit{Paris Research Centre} \\
\textit{Huawei Technologies}\\
Boulogne-Billancourt, France \\
fadhel.ayed@huawei.com}
}

\maketitle

\begin{abstract}
The increasing interest in Large Language Models (LLMs) within the telecommunications sector underscores their potential to revolutionize operational efficiency. However, the deployment of these sophisticated models is often hampered by their substantial size and computational demands, raising concerns about their viability in resource-constrained environments. Addressing this challenge, recent advancements have seen the emergence of \textit{small} language models that surprisingly exhibit performance comparable to their larger counterparts in many tasks, such as coding and common-sense
reasoning. \mbox{Phi-2}, a compact yet powerful model, exemplifies this new wave of efficient small language models. 
This paper conducts a comprehensive evaluation of \mbox{Phi-2}'s intrinsic understanding of the telecommunications domain. Recognizing the scale-related limitations, we enhance \mbox{Phi-2}'s capabilities through a Retrieval-Augmented Generation approach, meticulously integrating an extensive knowledge base specifically curated with telecom standard specifications. The enhanced \mbox{Phi-2} model demonstrates a profound improvement in accuracy, answering questions about telecom standards with a precision that closely rivals the more resource-intensive \mbox{GPT-3.5}. The paper further explores the refined capabilities of \mbox{Phi-2} in addressing problem-solving scenarios within the telecom sector, highlighting its potentials and limitations. 
\end{abstract}

%\begin{IEEEkeywords}
%component, formatting, style, styling, insert
%\end{IEEEkeywords}

\section{Introduction}
The increasing interest in \acp{LLM} has transcended beyond the boundaries of academia, leading to significant attention across various industry sectors. These sectors are not only investing in understanding the capabilities of \acp{LLM} but also actively exploring their potential to innovate and enhance operational processes. Although the transformative impacts of these models are yet to be fully realized, there is a concerted effort to harness the power of \acp{LLM} for potential groundbreaking changes across different fields.
%Impact of LLM in different domains
In the healthcare sector, for instance, LLMs such as GatorTronGPT have established new benchmarks in biomedical natural language processing~\cite{peng2023study}. In particular, their excellence in tasks such as relation extraction and question answering underscores their crucial role in decoding complex medical texts and supporting informed decision-making processes and healthcare quality~\cite{thirunavukarasu2023large}.
Similarly, the finance sector has begun to explore the capacity of LLMs to provide deep insights into market trends and to bolster risk analysis frameworks~\cite{wu2023bloomberggpt}.
These applications exemplify the broad potential of LLMs, setting the stage for the telecom sector, where the potential for \acp{LLM} is uniquely positioned. Telecom, with its vast generation and transmission of data, stands to benefit immensely from the advanced text comprehension and interaction capabilities of \acp{LLM}. 
%\textcolor{blue}{[Initial studies in telecom domain]}
The exploration of \acp{LLM} within the telecom sector has commenced with promising studies that pave the way for innovative applications. 
In~\cite{bariah2023understanding}, \acp{LLM} such as \ac{BERT} and \ac{GPT}-2 were leveraged to classify working groups within the \ac{3GPP} based on analysis of technical specifications.  Moreover, the potential of \acp{LLM} in facilitating complex Field-Programmable Gate Array (FPGA) development within wireless systems was highlighted in~\cite{du2023power}. Additionally, the authors in~\cite{bariah2023large} provided a vision of a future where \acp{LLM}, along with multi-modal data, can significantly contribute to the development of \ac{RAN} technologies such as beamforming and localization. In this future, by combining different data types like text and visuals, \acp{LLM} can potentially assist in optimizing and improving \ac{RAN} functionalities.
Finally, in~\cite{maatouk2023large} the short-term potentials of \acp{LLM} in telecom where analyzed and multiple use cases where depicted, from network troubleshooting to network modeling.
These initial studies collectively highlight the vast potential of \acp{LLM} in the telecom sector, setting a strong foundation for future research and practical implementations that could reshape the industry.
%Challenges related to big dimension of these models, energy consumption, capability of running them locally
Despite these promising applications, integrating \acp{LLM} into the telecom sector presents significant challenges, primarily due to their size and computational requirements. The high parameter count of these models necessitates significant computational resources for both training and operation, challenging their deployment in resource-constrained settings. 
Energy consumption is another critical concern, with the operation of these models requiring significant electrical power, thus increasing operational costs and environmental impact.
To provide a concrete example, the training of GPT-3 is estimated to have emitted 502 tonnes of CO2eq, which is 28 times the annual carbon footprint of an average American and 507 times the carbon footprint of a round trip flight from New York to San
Francisco~\cite{maslej2023artificial}.
%Raise of smaller models, like Phi, with much smaller dimensions, that has shown remarkable performance, sometimes at the level of much larger models

In light of the challenges associated with traditional \acp{LLM}, there is an increasing focus on developing compact language models that balance performance and efficiency. An exemplar of this trend is Phi-2, a \ac{SLM} recently released by Microsoft~\cite{gunasekar2023textbooks}
, addressing the size and energy issues prevalent in larger models. Phi-2's reduced size offers multiple advantages, including decreased computational resource requirements, lower energy consumption, lower environmental impact, and suitability for deployment in local environments like edge computing devices in telecom.

%In this paper, we compare a small model with SoTA LLMs in the telecom domain, etc
In this paper, we select Phi-2 as a reference \ac{SLM} and present a comprehensive analysis of its telecom knowledge, comparing its performance against larger models like \mbox{GPT-3.5} and GPT-4. Additionally, the paper investigates whether the performance of these smaller models can be significantly enhanced in certain domains by implementing a \ac{RAG} mechanism.
Finally, we delve into the capabilities and limitations of Phi-2 by examining its performance in two telecom-related tasks: the derivation of a mathematical model to estimate the energy consumption of a network, and the resolution of a user association problem. %These tasks provide a lens through which to evaluate the practical applicability and limitations of SLMs within the telecom sector.

\section{Methodology}
This section outlines the methodology employed to evaluate the telecom-specific knowledge of various language models, delineate the models under investigation, and introduce the concept of \ac{RAG} for enhancing the knowledge bases of language models.

\subsection{Evaluation of telecom knowledge}
This study utilizes the TeleQnA dataset, meticulously crafted and detailed in~\cite{maatouk2023teleqna}, to conduct a comparative analysis of language models, specifically Phi-2, GPT-3.5, and GPT-4, within the telecom domain.

The TeleQnA dataset, originating from an extensive corpus of telecom-specific documents, is structured into a series of \acp{MCQ}. Each \ac{MCQ}, comprising a question stem, a set of distractors, and a single correct answer, is intricately designed to probe a model's comprehension and application of telecom-related concepts. This structure not only facilitates a stringent evaluation of the models' performance but also mirrors the complexity and practical decision-making scenarios encountered in the telecom industry.

The dataset's content spans a broad spectrum of telecom topics, ensuring a comprehensive evaluation. In particular, questions are divided into 5 main categories, namely `Lexicon', `Research Overview', `Research Publications', `Standards Overview', and `Standards Specifications', encapsulating the breadth and depth of the domain, and ranging from foundational terminology to intricate protocol details and industry standards. This categorization enables a nuanced evaluation of LLMs' performance, facilitating the assessment of their potential applications and limitations within the telecom sector.

\subsection{Language models}
In this study, we undertake a comparative performance analysis, juxtaposing a \ac{SLM}, Phi-2, with its larger counterparts, GPT-3.5 and GPT-4. This section delineates the distinctive characteristics and architectural nuances of each model.
\subsubsection{GPT3.5}
GPT-3.5 emerges as a fine-tuned version of its predecessor, \mbox{GPT-3}, harboring 175 billion parameters. Developed with the specific intention of mitigating toxic outputs, GPT-3.5 has been a pivotal model in the transition from \mbox{GPT-3} to GPT-4. Notably, GPT-3.5's architecture includes 12 stacks of decoder blocks with multi-head attention, highlighting its intricate design aimed at elevating the standards of natural language processing. Introduced in January 2022, GPT-3.5 signifies a step forward in the evolution of the GPT series, integrating fine-tuning processes like \ac{RLHF} to refine its outputs based on human values, thereby enhancing model performance and steerability in a multitude of applications, including content creation, code writing, and customer service interactions. This capability to be fine-tuned and customized for specific use cases, coupled with the model's capacity to handle larger context windows (up to 16.000 tokens), underscores its adaptability and potential for efficient, tailored applications
%\textcolor{blue}{\textit{(https://iq.opengenus.org/gpt-3-5-model/, 
%https://www.iffort.com/blog/2023/03/31/gpt-3-vs-gpt-3-5/,
%https://openai.com/blog/gpt-3-5-turbo-fine-tuning-and-api-updates)}}

\subsubsection{GPT4}
GPT-4, the latest model in OpenAI's series of GPTs, is a multimodal LLM. While OpenAI has not disclosed the precise size of the model, it is rumored to have a staggering 1.76 trillion parameters, representing a monumental increase from its predecessors. 
GPT-4 showcases an ability to handle much more nuanced instructions than GPT-3.5, making it more reliable and creative. Notably, it offers a context windows of 128.000 tokens, significantly larger than the previous models, thus enabling a more comprehensive understanding and generation of content. 
On the performance front, GPT-4 demonstrates remarkable proficiency in academic and professional domains. It has been evaluated on a variety of standardized tests, scoring impressively across the board. For instance, it scored around the 90th percentile on the Uniform Bar Exam and the LSAT, and around the 94th percentile on the SAT~\cite{openai2023gpt4}. Importantly, these scores not only illustrate the model's wide-ranging knowledge but also its ability to understand and process complex, multifaceted information.
%\textcolor{blue}{\textit{(
%https://en.wikipedia.org/wiki/GPT-4, 
%https://hix.ai/hub/chatgpt/gpt-4-parameters, 
%https://ar5iv.org/abs/2303.08774)}}

\subsubsection{Phi-2}
Phi-2 is a recent entry in the suite of \acp{SLM} from Microsoft called “Phi” that achieve remarkable performance on a variety of benchmarks. With 2.7 billion parameters, Phi-2 demonstrates an impressive ability to match or even surpass models up to 25 times its size in terms of reasoning and language understanding. This model's training leverages a combination of high-quality synthetic and web datasets, honing its capabilities in common sense reasoning, language understanding, math, and coding. Notably, Phi-2 is not aligned through \ac{RLHF}, nor has it undergone instruction fine-tuning. However, it still exhibits reduced toxicity and bias compared to similar models~\cite{Microsoft2023Phi2}. The model's structure is based on the Transformer architecture, with a next-word prediction objective, and it took 14 days to train on 96 A100 \acp{GPU}.

\subsection{Retrieval-Augmented Generation}
\begin{figure*}
    \centering
    \includegraphics[width=16cm]{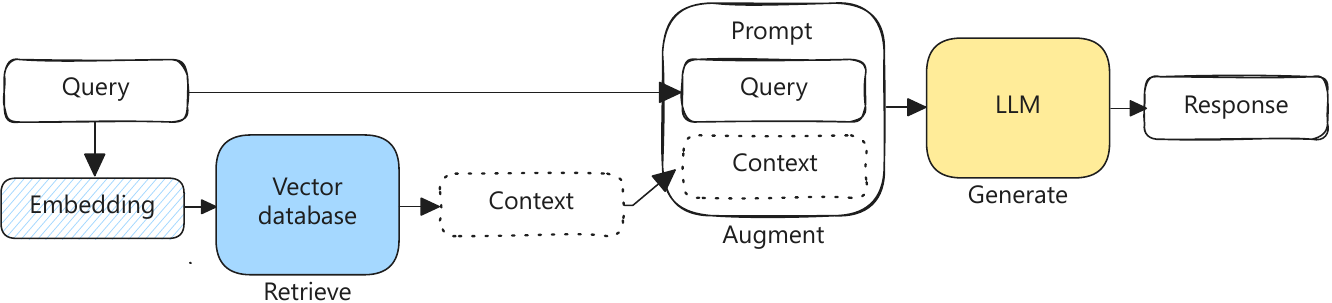}
    \caption{Scheme depicting the retrieve, augment and generate phases of the RAG mechanism}
    \label{fig:RAGscheme}
\end{figure*}
\ac{RAG} is a method that significantly enhances the output of \acp{LLM} by integrating an authoritative knowledge base external to the model's initial training data. This technique amplifies the intrinsic capabilities of \acp{LLM}, allowing them to remain pertinent, precise, and valuable across various subjects. 

At its core, \ac{RAG} addresses some of the intrinsic limitations of \acp{LLM}. While \acp{LLM} are proficient in generating responses based on a vast array of parameters representing general human language patterns, they fall short when detailed, up-to-date, or topic-specific responses are required. This is where \ac{RAG} steps in, providing a bridge to external, authoritative resources, and enriching the responses of \acp{LLM} with accurate information. 
Importantly, implementing \ac{RAG} is cost-effective compared to retraining the \ac{LLM} with new datasets.

The \ac{RAG} process starts with the creation of external data, often derived from various sources like databases and document repositories. This data is then converted into a format that is convenient for efficient retrieval, typically through embedding language models, and stored in a vector database to form a comprehensive knowledge library. Upon receiving a user query, the \ac{RAG} system performs a relevancy search, matching the query with the data in the vector database. This relevant information, along with the original user input, is then fed to the \ac{LLM}, which, equipped with this enriched context, generates a more accurate and detailed response. The process is illustrated in Figure~\ref{fig:RAGscheme}.

\section{Experimental Results}
In this section, we outline the experiments conducted to evaluate the knowledge capabilities of the \ac{SLM}, Phi-2, within the telecom domain. We compare the achieved scores of Phi-2 with those of significantly larger \acp{LLM}. Additionally, we investigate the potential for expanding Phi-2's knowledge base through the application of \ac{RAG}. Finally, we present two straightforward use cases. In the first use case, we demonstrate the enhanced capabilities of Phi-2 when empowered with \ac{RAG} of performing a modeling task within the telecom domain. In the second use case, we assess the capability of Phi-2 in solving an intricate user association problem.

\subsection{Assement of the telecom knowledge}
\begin{table}
\centering
\caption{Comparison of accuracies for different models}
\begin{tabular}{lcccc}
\hline
\textbf{Category} & \textbf{GPT-3.5 } & \textbf{GPT-4.0} & \textbf{Phi-2} \\ \hline
Lexicon (500) & 82.20\% & 86.80\% & 52.60\% \\
Research overview (2000) & 68.50\% & 76.25\% & 58.38\% \\
Research publications (4500) & 70.42\% & 77.62\% & 54.14\% \\
Standard overview (1000) & 64.00\% & 74.40\% & 48.04\% \\
Standard specifications (2000) & 56.97\% & 64.78\% & 44.27\% \\
\hline
\textbf{Overall accuracy (10000)} & 67.29\% & 74.91\% & 52.30\% \\
\hline
\end{tabular}
\label{table:comparison}
\end{table}
The TeleQnA dataset was adopted to assess the telecom knowledge of the Phi-2 model and compare the achieved score with those of GPT-3.5 and GPT-4.
To perform the knowledge evaluation, the Phi-2 model was presented with the following structured prompt:
\begin{tcolorbox}[colback=lightgray!20, colframe=lightgray!50,
boxrule=1pt,leftrule=5pt,arc=0pt,auto outer arc]
\vspace{-0.3cm}
\begin{lstlisting}[breaklines, basicstyle=\small\ttfamily]
Instruct: Answer the following question. 
Your answer must start with the number 
of the correct answer followed by the 
text of the answer.
(*@\textit{\{question\}}@*)
1. (*@\textit{\{answer 1\}}@*)
2. (*@\textit{\{answer 2\}}@*)
...
n. (*@\textit{\{answer n\}}@*)
Output:
\end{lstlisting}
\vspace{-0.3cm}
\end{tcolorbox}
In this prompt, placeholders \textit{\{question\}} and \textit{\{answer 1...n\}} were replaced with the respective questions and answer choices from each dataset entry.

We conducted a systematic evaluation by iteratively presenting the 10,000 questions available in the TeleQnA dataset to the Phi-2 model. The outputted responses were parsed to extract the selected option number, which was then compared to the correct choice indicated in the dataset. The overall evaluation process demanded 13 hours of computational resources on an NVIDIA V100 GPU.

Table~\ref{table:comparison} illustrates the performance achieved by the Phi-2 model, alongside the accuracy attained by the larger models, GPT-3.5 and GPT-4.

GPT-4 stands out as the top performer, with an overall accuracy of 74.91. This model excelled across all categories, particularly in `Lexicon' and `Research Publications', indicating its good understanding of telecom-related knowledge. In comparison, GPT-3.5 achived an overall accuracy of 67.29. While it trailed behind GPT-4, its performance, especially in `Lexicon' and `Research Publications', underscores its effective grasp of the domain.
%Surprisingly, Phi-2, despite its smaller architecture, achieved a noteworthy overall accuracy of 52.30.
Surprisingly, Phi-2 achieved a noteworthy overall accuracy of 52.30, despite its compact size, being 65 times smaller than GPT-3.5 and 652 times smaller than GPT-4.
Notable strengths lie in `Research Overview' and `Research Publications'. This accomplishment underscores the effectiveness of smaller models in specialized domains, even though they may have certain limitations when compared to their larger counterparts.

Importantly, all models encountered their greatest challenge in the `Standard Specifications' category. Telecom standards, governed by organizations such as \ac{3GPP}, \ac{IEEE}, and \ac{ITU}, are pivotal for ensuring compatibility and representing the intricate nature of telecommunications.
Despite the sophistication of the tested language models, this category posed distinctive challenges, characterized by intricate technical language and detailed specifications.
However, given the significance of this domain, it's imperative for language models to possess a strong foundation in telecom standards. 

%To tackle this issue, the next section delves into \ac{RAG} as a potential solution for enhancing language performance in this specialized domain.

\subsection{Improving knowledge of standard specifications}
\begin{figure}
    \centering
    \begin{tikzpicture}
    \begin{axis}[
        width=8cm,
        height=6cm,
        ybar,
        enlarge x limits=0.15,
        legend style={at={(0.5,-0.15)},
        anchor=north,legend columns=-1},
        ylabel={Accuracy (\%)},
        symbolic x coords={GPT-3.5,GPT-4,Phi-2,Phi-2+RAG},
        xtick=data,
        nodes near coords,
        nodes near coords align={vertical},
        ymin=0,ymax=75,
        ylabel near ticks,
        xlabel near ticks,
    ]
    \addplot [fill=nicolasred] coordinates {(GPT-3.5,56.97) (GPT-4,64.78) (Phi-2,44.27) (Phi-2+RAG,56.63)};
    \end{axis}
    \end{tikzpicture}
    \caption{Accuracy achieved by different language models over the `Standards Specifications' category.}
    \label{fig:models_standard_performance}
\end{figure}
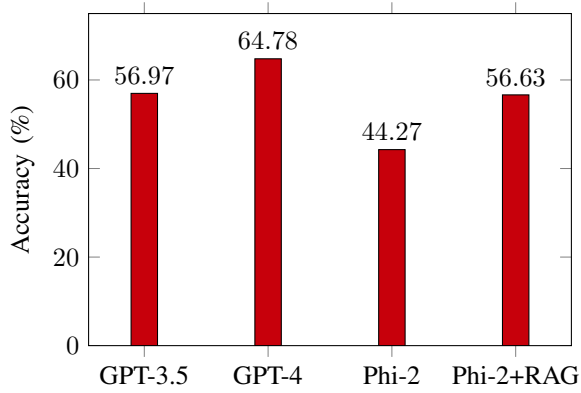

To enhance Phi-2's proficiency in the `Standards Specifications' category, we employed the \ac{RAG} technique. This involved a process of sourcing, processing, and embedding technical documents from recognized standards bodies, specifically \ac{3GPP}. The documents, initially in DOC/DOCX and PDF formats, were uniformly converted to TXT files to standardize the input for subsequent processing.

In the embedding phase, we utilized the bge-base-en-v1.5 model~\cite{bge_embedding}, employing a chunk size of 512 tokens. This chunking strategy was crucial for minimizing noise and maintaining semantic relevance within the content. The rationale behind this is rooted in the principles of semantic search: each segment, or `chunk,' encapsulates concentrated information pertinent to a particular topic. Proper chunking ensures that the search mechanism can precisely align with the essence of the query. Inappropriate chunk sizes could result in either overly broad results that miss specific details or overly narrow results that overlook contextually relevant information.

The resultant embeddings were systematically stored in a vector database and indexed to facilitate the retrieval during the query phase. In this phase, the questions and their corresponding multiple-choice options were used to navigate the database, aiming to retrieve the most significant embeddings. These embeddings were then provided as contextual data to the Phi-2 model, with the intent to bolster its capacity to accurately respond to questions.

The impact of integrating the RAG mechanism into \mbox{Phi-2}'s processing pipeline was profound. Figure~\ref{fig:models_standard_performance} illustrates a comparative analysis of the model's performance in addressing questions from the `Standards Specifications' category. The implementation of \ac{RAG} led to a notable increase in \mbox{Phi-2}'s accuracy, from 44.27\% to 56.63\%. This enhancement not only signifies the effectiveness of the RAG method in enriching the model's contextual understanding but also positions \mbox{Phi-2}'s performance in close proximity to that of GPT-3.5, which achieved an accuracy of 56.97\%. The convergence of \mbox{Phi-2}'s performance with that of a more advanced model like \mbox{GPT-3.5} underscores a crucial narrative in AI development: the thoughtful integration of advanced methodologies like RAG can significantly enhance the capabilities of compact models, enabling them to operate at a level traditionally reserved for their more complex counterparts.

\subsection{Use case: Network modeling}
\begin{figure}
    \centering
    \begin{tikzpicture}[]
    \begin{axis}[xlabel={BS load}, ylabel={Normalized energy consumption},width=8cm,height=6cm, %legend style={anchor=north east,draw=rounded corners},legend cell align={left}
    legend style={
        anchor=north east,
        legend cell align={left},
        % Increase inner separation for x and y
        inner xsep=10pt, % Adjust the horizontal padding
        inner ysep=3pt, % Adjust the vertical padding
        % Round the corners of the legend box
        rounded corners,
        % Optionally, specify the radius for rounded corners
        /tikz/rounded corners=1pt
    }
    ]
    \addplot[only marks, mark size=2pt, mark options={fill=gray,draw opacity=0, fill opacity=0.8}] table [x=Load, y=Power, col sep=comma] {phi2plot.csv};
    \addlegendentry{Ground truth}
    \addplot[only marks, mark size=2pt, mark options={fill=nicolascyan,draw opacity=0}] table [x=Load, y=phi2, col sep=comma] {phi2plot.csv};
    \addlegendentry{Phi-2}
    \addplot[only marks, mark size=2pt, mark options={fill=nicolasred,draw opacity=0}] table [x=Load, y=phi2rag, col sep=comma] {phi2plot.csv};
    \addlegendentry{Phi2+RAG}
    \end{axis}
    \end{tikzpicture}
    \caption{Ground-truth energy consumption at different BS load levels and estimation performed by Phi-2 standalone and Phi-2 with RAG.}
    \label{fig:network_results}
\end{figure}
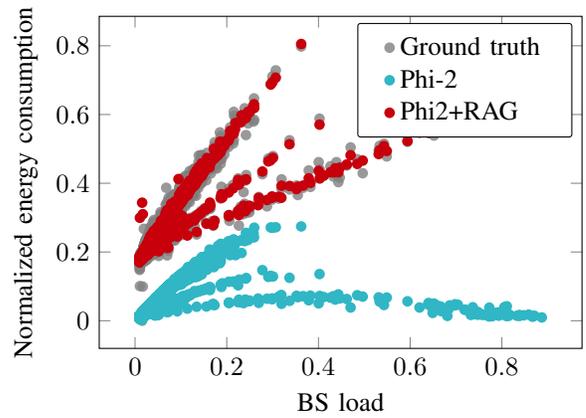
In this section, we investigate a use case focused on modelling the energy consumption within a telecommunications network. This analysis specifically targets a simplified network architecture consisting of 90 single-carrier \acp{BS} employing a single energy-saving method (i.e., symbol shutdown).

The Phi-2 model, both in its standalone and RAG-augmented versions, was tasked with a mathematical model construction exercise. The  \ac{SLM} had access to a set of data features encompassing variables such as BS location, carrier frequency, \ac{BS} load, and more. The primary objective was to discern the critical parameters for estimating the energy consumption of a \ac{BS} (task 1) and to formulate a mathematical model that encapsulates the relationship between the selected parameters and the overall energy consumption (task 2).

For task 1,  the models were instructed to identify essential parameters for energy consumption estimation from a designated list through the following prompt:
\begin{tcolorbox}[
    enhanced, 
    breakable,
    skin first=enhanced,
    skin middle=enhanced,
    skin last=enhanced,
    colback=lightgray!20, colframe=lightgray!50,boxrule=1pt,leftrule=5pt,arc=0pt,auto outer arc]
\vspace{-0.3cm}
\begin{lstlisting}[breaklines,basicstyle=\small\ttfamily]
Instruct: Select, based on your 
knowledge, the most important parameters 
for estimating the energy consumption of 
a mobile base station in a mathematical 
model. Select the relevant parameters 
from the list:
- BS load
- latitude
- longitude
- serial number
- production year
- maximum transmit power
- duration of activation of symbol shutdown
- weight
- number of antennas
Output:
\end{lstlisting}
\vspace{-0.3cm}
\end{tcolorbox}

Both the Phi-2 and Phi-2+RAG models identified BS load ($L$), maximum transmit power ($MTX$), and duration of activation of symbol shutdown ($DSS$) as the key influencers of the energy consumption.

Task 2 consisted in deriving a mathematical model for estimating energy consumption using the parameters identified in the preceding task. To accomplish this, both models were presented with the following prompt:
\begin{tcolorbox}[colback=lightgray!20, colframe=lightgray!50,
boxrule=1pt,leftrule=5pt,arc=0pt,auto outer arc]
\vspace{-0.3cm}
\begin{lstlisting}[breaklines, basicstyle=\small\ttfamily]
Instruct: write a mathematical formula 
to estimate the energy consumption of a 
base station using the following 
parameters:
- BS load (L)
- maximum transmit power (MTX)
- duration of activation of symbol shutdown (DSS)
Output:
\end{lstlisting}
\vspace{-0.3cm}
\end{tcolorbox}

The Phi-2 model provided an energy consumption model represented by the equation:
\begin{equation}
    E=L\cdot MTX \cdot DSS
\end{equation}
%However, this formulation is fundamentally flawed.% and did not align well with the actual energy consumption patterns of the network, yielding a high \ac{MAPE} of 78\%.

In contrast, the Phi-2+RAG model, leveraging its enriched knowledge base filled with documents on the energy consumption behaviors of typical \acp{BS}, suggested a more sophisticated and accurate formula for energy consumption:
\begin{equation}
    E=PS-\alpha DSS +1/ \epsilon \cdot L\cdot MTX
\end{equation}
Within this formula, as elucidated by the Phi-2+RAG's output, $\alpha$ signifies a scaling factor, PS represents static energy consumption, and $\epsilon$ denotes the efficiency of the power amplifier.

Figure~\ref{fig:network_results} illustrates the ground-truth energy consumption values corresponding to different BS load levels, in conjunction with the estimations provided by two distinct models.

Upon fitting the two model with the real network data, the Phi-2 model demonstrated to be not aligned with the actual energy consumption patterns of the network, yielding a high \ac{MAPE} of 78\%.
Conversely, Phi-2+RAG model demonstrated a remarkable ability to capture the actual energy consumption trend, reflected in a significantly lower \ac{MAPE} of 4.3\%.

This comparative analysis highlights the efficacy of the \ac{RAG} methodology in refining the modelling capabilities of \acp{SLM} like Phi-2. By integrating targeted, domain-specific knowledge, Phi-2+RAG not only narrowed the gap in performance relative to larger models but also exhibited an enhanced understanding of intricate energy dynamics in telecom networks. This underscores the potential of \ac{RAG} as a powerful tool for augmenting the knowledge base of language models, thereby driving superior performance in specialized tasks.

\subsection{Use case: User association problem}
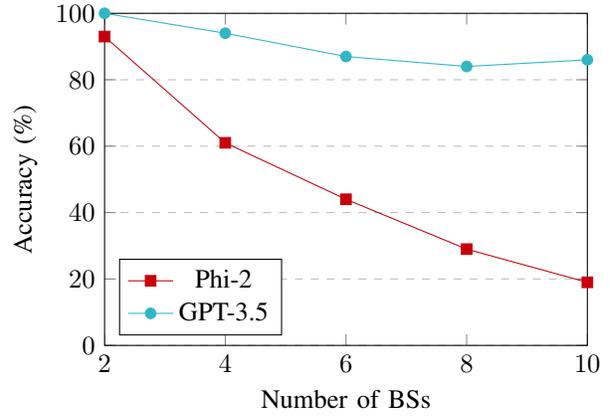
\begin{figure}
\begin{tikzpicture}
\begin{axis}[
    width=8cm,
    height=6cm,
    xlabel={Number of BSs},
    ylabel={Accuracy (\%)},
    xmin=2, xmax=10,
    ymin=0, ymax=100,
    xtick={2,4,6,8,10},
    ytick={0,20,40,60,80,100},
    legend pos=south west,
    ymajorgrids=true,
    grid style=dashed,
]

\addplot[
    color=nicolasred,
    mark=square*,
    ]
    coordinates {
    (2,93)(4,61)(6,44)(8,29)(10,19)
    };
    \addlegendentry{Phi-2}

\addplot[
    color=nicolascyan,
    mark=*, 
    ]
    coordinates {
    (2,100)(4,94)(6,87)(8,84)(10,86)
    };
    \addlegendentry{GPT-3.5}

\end{axis}
\end{tikzpicture}
\caption{Accuracy achived by GPT-3.5 and Phi-2 in resolving the user association problem.}
\label{fig:problem_resukts}
\end{figure}

In this section, we evaluate the proficiency of the \mbox{Phi-2} model in addressing complex problem-solving scenarios necessitating nuanced logical reasoning. This assessment is premised on a series of problems where Phi-2 is tasked with determining the optimal \ac{BS} connections for a \ac{UE} based on varying signal strength readings and a condition to systematically avoid the strongest \ac{BS}. An illustrative problem is presented as follows:

\begin{tcolorbox}[
    colback=lightgray!20, colframe=lightgray!50,boxrule=1pt,leftrule=5pt,arc=0pt,auto outer arc]
\vspace{-0.3cm}
\begin{lstlisting}[breaklines,basicstyle=\small\ttfamily]
Instruct: A mobile device receives 
signals from three different base 
stations. The signal strengths are as 
follows:
- The signal strength from base station 1 is -80 dBm
- The signal strength from base station 2 is -62 dBm
- The signal strength from base station 3 is -70 dBm
The device must connect to the base 
station providing the strongest signal 
but avoiding base station 2.
Given these signal strengths, to which 
base station should the mobile device 
connect? 
Output:
\end{lstlisting}
\vspace{-0.3cm}
\end{tcolorbox}

The task necessitates the language model to interpret the directive to connect to the strongest signal, identify the strongest signal amidst the options, recognize the instruction to disregard the strongest signal, and subsequently select the second strongest signal for connection.

A diverse array of problems, encompassing various signal strength values and numbers of BSs to select, was formulated and deployed to evaluate the model's performance. Figure \ref{fig:problem_resukts} illustrates the precision achieved by Phi-2 in scenarios with different numbers of BS. Specifically, Phi-2 exhibits a 93\% accuracy rate in scenarios with two options, with observed accuracy decreasing linearly as the number of potential choices increases. In contrast, a larger model, such as GPT-3.5, consistently maintains an accuracy rate exceeding 85\%, regardless of the number of available options.

These findings highlight a crucial limitation inherent to \acp{SLM} like Phi-2: while these models excel at efficiently retrieving information and executing tasks predicated on this retrieval, they encounter difficulties in scenarios that demand elaborate and multistep reasoning. This limitation stems partly from the inherent compact design and training of these models, which may not fully encompass the depth of logical reasoning processes achievable by larger models.
Addressing this limitation of \acp{SLM} may be partially possible by leveraging advancements in language model reasoning capabilities, such as the technique of chain-of-thought prompting, which has been identified as a breakthrough in enabling language models to decompose multi-step problems into intermediate steps, thereby improving their ability to tackle complex reasoning challenges~\cite{wei2022chain}.
Additionally, specializing \acp{SLM} towards multi-step reasoning through methods such as distillation and data augmentation has shown promise. This approach involves fine-tuning these models on specific reasoning tasks using data generated from larger models or specialized training datasets, enhancing their reasoning capabilities despite potential trade-offs between general and specialized skills~\cite{fu2023specializing}.
\balance
\section{Conclusions}
This paper presented a comprehensive analysis of the capabilities and limitations of SLMs, with a focus on Phi-2, in the context of the telecommunications domain. Through systematic evaluation against larger counterparts like GPT-3.5 and GPT-4, this study underscored the potential of SLMs to operate efficiently within specific niche sectors, despite their relatively compact size and lower computational demands.

Our findings revealed that while SLMs like Phi-2 exhibit commendable performance across a range of tasks, their proficiency significantly benefits from augmentation techniques such as RAG. The integration of RAG, particularly in complex problem-solving scenarios and knowledge-intensive tasks, markedly enhances the accuracy and applicability of SLMs, bridging the gap between them and their larger counterparts.

The experimental results demonstrated that Phi-2, when enhanced with a specialized knowledge base through RAG, achieves performance levels comparable to GPT-3.5 in tasks requiring detailed understanding of telecom standards specifications. This is a notable advancement, highlighting the effectiveness of leveraging external databases to supplement the intrinsic knowledge of SLMs. Furthermore, the use case of network modeling illustrated how RAG could empower Phi-2 to formulate more accurate and realistic models for estimating energy consumption in telecom networks, showcasing the practical utility of SLMs in operational telecom environments.

Finally, the user association problem use case evaluated \mbox{Phi-2}'s logical reasoning abilities in complex scenarios, revealing inherent limitations in handling tasks that require intricate reasoning and multi-step thinking. 
In particular, the consistent performance of larger models in these scenarios emphasizes the trade-off between model size, computational efficiency, and task complexity.

%\begin{figure}
%    \centering
%    \includegraphics[width=8cm]{energyPhi.pdf}
%    \caption{Ground-truth energy consumption at different BS load levels and estimation performed by Phi-2 standalone and Phi-2 with RAG.}
%    \label{fig:network_results}
%\end{figure}

\begin{acronym}[AAAAAAAAA]
 \acro{3GPP}{third generation partnership project}
 \acro{AI}{artificial intelligence}
 \acro{API}{application programming interface}
 \acro{BS}{base station}
 \acro{BERT}{bidirectional encoder representations from transformers}
 \acro{FPGA}{field-programmable gate array}
 \acro{GPT}{generative pre-trained transformer}
 \acro{IEEE}{institute of electrical and electronics engineers}
 \acro{ITU}{international telecommunication union}
 \acro{GPU}{graphical processing unit}
 \acro{LLM}{large language model}
 \acro{MAPE}{mean absolute percentage error}
 \acro{MCQ}{multiple-choice questions}
 \acro{RAG}{retrieval-augmented generation}
 \acro{RAN}{radio access network}
 \acro{RLHF}{reinforcement learning with human feedback}
 \acro{SLM}{small language model}
 \acro{UE}{user equipment}
 \end{acronym}

\bibliographystyle{IEEEtran}
\bibliography{reference.bib}

\end{document}